%% file: main.tex
\documentclass{article}
\usepackage{spconf,amsmath,graphicx}
\usepackage{amsmath, bm}
\usepackage{mathrsfs} 
\usepackage{amsfonts}
\usepackage{amsthm}
\usepackage{enumitem}
\usepackage{hyperref}       % hyperlinks
\usepackage{macros}
\usepackage{subcaption}

\usepackage{cite}

\title{Space-Time Graph Neural Networks with Stochastic Graph Perturbations}
\name{Samar Hadou, Charilaos I. Kanatsoulis, and Alejandro Ribeiro}
\address{Department of Electrical and Systems Engineering, University of Pennsylvania, PA, USA}
\begin{document}
\ninept
\maketitle
\begin{abstract}
\input{Sections/Abstarct}
\end{abstract}
\begin{keywords}
ST-GNNs, GNNs, stability anaysis, graph perturbations, transfer learning
\end{keywords}
\section{Introduction}
\label{sec:intro}
\input{Sections/Intro}

\section{Space-Time Graph Neural Networks}
\label{sec:STGNNs}
\input{Sections/STGNNs}
\section{Generalized Space-Time GNNs}
\label{sec:generalized}
\input{Sections/Stochastic}
\section{Stability To Stochastic Graph Perturbations}
\label{sec:stability}
\input{Sections/Stability}

\section{Numerical Simulations}
\label{sec:sim}
\input{Sections/Simulations}

\section{Conclusions}
\label{sec:con}
\input{Sections/Conclusions}

\bibliographystyle{IEEEbib}
\bibliography{myBib}

\end{document}

%% file: Sections/Abstarct.tex
Space-time graph neural networks (ST-GNNs) are recently developed architectures that learn efficient graph representations of time-varying data. ST-GNNs are particularly useful in multi-agent systems, due to their stability properties and their ability to respect communication delays between the agents. In this paper we revisit the stability properties of ST-GNNs and prove that they are stable to stochastic graph perturbations. Our analysis suggests that ST-GNNs are suitable for transfer learning on time-varying graphs and enables the design of generalized convolutional architectures that jointly process time-varying graphs and time-varying signals. Numerical experiments on decentralized control systems validate our theoretical results and showcase the benefits of traditional and generalized ST-GNN architectures.

%% file: Sections/Intro.tex
Stability analysis of graph neural networks (GNNs) was studied extensively in literature
\cite{verma2019, keriven2020, Gama2020, nguyen2022, kenlay2021} and affirms that small changes in input graphs translate into small, bounded changes in the outputs. It also provides a characterization of the design parameters, e.g., the depth and width of GNNs, that preserve stability. The acquired gain, when stability holds, is the ability to transfer learned GNNs between different graphs of the same size without re-training the models.
Stability is a notion that is different from GNN transferability, which considers transfer learning to graphs of larger sizes \cite{levie2021, ruiz2020, maskey2021}. Together, they allow executing GNNs on underlying graphs that are different from those used in training and determine which architectures are more robust to graph changes.

GNNs are aligned with applications in distributed controllers \cite{osanlou2019, li2021, zhou2022} and wireless systems \cite{he2021,zhao2021,yan2021}. Both tasks are challenging, since the underlying graph varies rapidly over time and the signals endure delays while being propagated between agents. To deal with signal delays, we leveraged diffusion equations to introduce space-time graph neural networks (ST-GNNs) in \cite{hadou2021}. Specifically, we formulated a diffusion equation over a space-time domain that respects propagation time between nodes. This allowed the proposed space-time convolutions to aggregate each node's present data along with outdated information from its neighbors. During training ST-GNNs learn to weigh the outdated information and offset their negative impact on the model's predictions.
Our analysis also proved that the new architecture is stable to graph and time perturbations. However, the challenge of facing dynamic graphs during training is still to be addressed.

The first work to approach this challenge is \cite{GAO2021} that studied dynamic graphs modelled as stochastic processes and showed that GNNs are stable to stochastic graph perturbations. However, their analysis focuses on signals that are fixed over time, which is impractical in certain applications. 
In this paper, we bridge this gap and study both signals and graphs that vary over time. We extend our previous work in ST-GNNs to accommodate time-varying graphs and prove their stability to stochastic graph perturbations.
Our contributions can be summarized as follows:
\begin{enumerate}[label=\textbf{(C\arabic*)}]
    \item We prove the stability of STGFs and ST-GNNs to stochastic graph perturbations. Our result implies that ST-GNNs can handle transfer learning to time-varying graphs. 
    % Our theorems also characterize the stability bound and show that deep GNNs are less robust to graph perturbations.
    \item Motivated by our stability analysis, we introduce generalized STGFs and generalized ST-GNNs that are tailored to jointly process time-varying graphs and time-varying signals and study their spectral properties.
    % \item  We study the generalized ST-GNNs in the frequency domain and and show their multi-variate frequency response.
    \item We conduct experimental examination of ST-GNNs and generalized ST-GNNs distributed controller settings.
\end{enumerate}
 Due to the limited space, all the proofs are relegated to an extended version of the paper.

%% file: Sections/STGNNs.tex
Consider an undirected graph $\cG = (\cV, \cE)$ with a set of $N$ nodes $\cV$ and a set of edges $\cE \subseteq \cV \times \cV$. 
We define discrete-time signals ${\bf x}_i(t)$ at each node $i$, which can be concatenated to a space-time graph signal ${\bf X} \in \mathbb{R}^{N \times T}$. Then $\bf S\bf X$ and $\bf X\bf C$ define the 1-hop diffusion of the space-time signal in space and time respectively. Note that ${\bf S}\in \mathbb{R}^{N\times N}$ is the graph shift operator (GSO), which usually represents the graph adjacency or Laplacian, and ${\bf C}\in \{0,1\}^{N\times N}$ is a circulant matrix that models the time shift operator (TSO). Then we can define the $k$-hop space-time diffusion as ${\bf S}^k {\bf X} {\bf C}^{k}$, which allows us to define the space-time graph filter (STGF) as the following convolutional operator 
\begin{equation} \label{eq:STGF}
    {{ \bf Y}} = \sum_{k= 0}^K h_k {\bf S}^k {\bf X} {\bf C}^{k},
\end{equation}
where the coefficients $\{h_k\}_{k=0}^K$ and the number of filter taps $K$ are design parameters. The operation in \eqref{eq:STGF} respects the time delays that occur in sharing information between the neighbors. Indeed, the GSO and TSO are jointly applied to ensure that the signals are shifted in space and time before being aggregated over the graphs. 

We design ST-GNNs as a cascade layers that deploy the STGF in \eqref{eq:STGF} followed by a nonlinear activation function $\sigma$. To distinguish the layers' intermediate outputs we use subscript $l$ and superscript $f$ to denote the $l$-th layer and the $f$-th feature of the signal respectively. The output of the $l$-th layer is expressed as
\begin{equation} \label{eq:STGNN}
    {\bf X}_l^f = \s \Big( \sum_{g = 1}^{F}\sum_{k= 0}^K h_{kl}^{fg} {\bf S}^k {\bf X}_{l-1}^g {\bf C}^{k}\Big).
\end{equation}
 We assume that the number of the features $F$ is fixed across all layers. We further refer to the ST-GNN's output, i.e., the output of the last layer $L$, as ${\bf \Phi}({\bf X}; {\bf S}, \cH)$ with $\cH = \{h_{kl}^{fg}\}_{k,l}^{f,g}$ being the set of all learnable parameters. Note that, we represent the underlying graph with a fixed GSO $\bf S$, which is not always the case in practise. In the following section, overcome this limitation we show how \eqref{eq:STGF} can be modified to accommodate time-varying graphs.

%% file: Sections/Stochastic.tex
In various applications the graph is changing over time. In multi-agent systems, for instance, the graph dynamics are time varying due to link dropping and re-wiring. As a result, we observe a sequence of GSOs $\{ {\bf S}_k, \dots, {\bf S}_0\}$. The diffusion of graph signals over this sequence is modeled as \[{\bf S}_k \dots {\bf S}_0 {\bf X}=: {\bf S}_{k:0}{\bf X}.\] Since STGFs perform a weighted sum of diffused signals as shown in \eqref{eq:STGF}, a generalized form of the STGF's output can be derived as
\begin{equation} \label{eq:stochSTGF}
    \tilde{\bf Y} = \sum_{k= 0}^K h_k {\bf S}_k \dots {\bf S}_0 {\bf X C}^k = \sum_{k= 0}^K h_k {\bf S}_{k:0} {\bf X C}^k,
\end{equation}
with ${\bf S}_0$ being the identity matrix ${\bf I}$.
Equation \eqref{eq:STGF} is obviously a special case with ${\bf S}_K =\dots ={\bf S}_1={\bf S}$.
Like ST-GNNs in \eqref{eq:STGNN}, we define the output of the $l$-th layer of the generalized ST-GNN as
\begin{equation} \label{eq:gSTGNN}
    \tilde{\bf X}_l^f = \s \Big( \sum_{g = 1}^{F}\sum_{k= 0}^K h_{kl}^{fg} {\bf S}_{k:0} \tilde{\bf X}_{l-1}^g {\bf C}^{k}\Big).
\end{equation}

\noindent The following lemma shows the frequency response of \eqref{eq:stochSTGF}.

\begin{lemma*}[Generalized STGF frequency response] \label{def:generalizedFR}
For a space-time graph filter with coefficients $\{h_k\}_{k=0}^K$ running over a sequence of $K$ GSOs as in \eqref{eq:stochSTGF}, the generalized frequency response is
\begin{equation} \label{eq:FRstochastic}
    \begin{split}
        h({\bm \l}, \om) = \sum_{k=0}^K h_k e^{jk\om} \prod_{\k=0}^k \l_{\k}, 
    \end{split}
\end{equation}
where ${\bm \lambda} = [\lambda_1, \dots, \lambda_K]^\top$, $\lambda_0 = 1$, and $\lambda_k$ is the analytic variable corresponding to the eigenvalue of ${\bf S}_k$.
\end{lemma*}
The proof of Lemma \ref{def:generalizedFR} is delegated to the journal version, due to space limitations. We observe that the frequency response is a multi-variate function in the graph-frequency variables $\l_k, \forall k$, and the time-frequency variable $\om$. 

In practice, we usually need to transfer generalized ST-GNN to different time-varying graphs that are not observed during training. Therefor a natural question that arises is ``under which conditions is transfer learning doable?'' We tackle give a definitive answer to this question by analyzing the deviation in the outputs between ST-GNNs and generalized ST-GNNs. Bounded deviations, imply that learned models can be transferred between different sequences of graphs.

%% file: Sections/Stability.tex
To analyze the deviation between ST-GNNs and genaralized ST-GNNs, we first define a model for time-varying graphs. In particular, we study a sequence of random graphs $\{{\cal G}_k\}_{k=1}^K$ sampled from a nominal graph $\cal G$ according to the following criteria.
\begin{definition*} [Random Edge Sampling] \label{def:RES}
Given a graph $\cG = (\cV, \cE )$, we define a random graph RES($\cG, p$) with a realization $\cG_k = (\cV, \cE_k)$ such that an edge $(i, j) \in \cE$ exists in $\cE_k$ with probability $p$, i.e.,
\begin{equation}
    \mathbb{P}[(i, j) \in \cE_k] = p, \ \forall (i, j)\in \cE.
\end{equation}
\end{definition*} 
Definition \ref{def:RES} implies that ${\cal G}_k$  is constructed by sampling edges from the nominal graph with probability $p$. From Definition \ref{def:RES}, it follows that the relation between the nominal GSO $\bf S$ and the random GSO ${\bf S}_k$ can be written as
\begin{equation} \label{eq:perturbationModel} 
    {\bf S}_k = {\bf S} + {\bf E}_k,
\end{equation}
where ${\bf E}_k$ models the deviation between the two GSOs. In \eqref{eq:perturbationModel}, The size of the deviation ${\bf E}_k$ is controlled by the edge-drop probability $1-p$. Higher probabilities account for dropping more edges, which in turn results in higher perturbation size $\|{\bf E}_k\|_2$. Probability $p$ can be viewed as a measure of similarity between the nominal graph ${\cal G}$ and a random realization ${\cal G}_k$. 

With relation \eqref{eq:perturbationModel} in place, we next aim to compute the difference in STGF's outputs when running over either $\bf S$ or the sequence ${\bf S}_{K:0}$ with each GSO having been constructed according to Definition \ref{def:RES}.
To achieve this, we introduce Lemma \eqref{lem:LipGradient}, which defines a smoothness measure of the multi-variate filter in \eqref{eq:FRstochastic}. 

\begin{lemma*}\label{lem:LipGradient}
Consider the generalized frequency response $h({\bm \l}, \om)$ in Lemma \ref{def:generalizedFR}, and
let ${\bm \l}_1 = [\l_{11}, \dots, \l_{1K}]^T$ and ${\bm \l}_2 = [\l_{21}, \dots, \l_{2K}]^T$ be two multivariate graph-frequency vectors. The variability in the frequency response is then quantified as
\begin{equation}
    \label{eq:grad}
    h({\bm \l}_1, \om) - h({\bm \l}_2, \om) = {\bf \nabla}_{\bm \l}^T h({\bm \l}_{1,2}, \om) \cdot ({\bm \l}_1 - {\bm \l}_2),
\end{equation}
where ${\bf \nabla}_{\bm \l} h({\bm \l}_{1,2}, \om) = \left[\frac{\partial h({\bm\l}^{(1)}, \om)}{\partial \l_1}, \dots, \frac{\partial h({\bm\l}^{(K)}, \om)}{\partial \l_K} \right]^T$ is the Lipschitz gradient, and ${\bm\l}^{(k)} = [\l_{11}, \dots, \l_{1k},\allowbreak \l_{2(k+1)}, \dots, \l_{2K}]^T$ contains the first $k$ entries of ${\bm \l}_1$ along with the last $K-k$ entries of ${\bm \l}_2$.
\end{lemma*}
The proof of Lemma \ref{lem:LipGradient} follows the proof of Lemma 1 in \cite{GAO2021}.
The Lipschitz gradient in \eqref{eq:grad} can be seen as a general measure of filter-response variability, which is usually quantified by the derivatives in univariate filters. Using the Lipschitz gradient, we define a class of smooth filters, namely generalized integral Lipschitz filter, in the following definition.

\begin{assumption*} \label{ass:Lipschitz} [Generalized Integral Lipschitz filter] An STGF with the generalized frequency response in Lemma \ref{def:generalizedFR} is generalized integral Lipschitz if there exists a constant $C_L > 0$ such that for any graph-frequency vectors ${\bm \l}_1$ and ${\bm \l}_2$ it holds that for all $\om$
\begin{equation}\label{eq:Lipschitz}
\begin{split}
    \| {\bf \nabla}_{\bm \l} h({\bm \l}_{1,2}, \om)\|_2 \leq C_L,
\end{split}
\end{equation}
\begin{equation}\label{eq:Lipschitz2}
    \quad \|{\bm \l}_1 \odot {\bf \nabla}_{\bm \l} h({\bm \l}_{1,2}, \om) \|_2 \leq C_L,
\end{equation}
where ${\bf \nabla}_{\bm \l} h({\bm \l}_{1,2}, \om)$ is the Lipschitz gradient defined in Lemma \ref{lem:LipGradient}, and $\odot$ is the Hadamard product.
\end{assumption*}
The condition in \eqref{eq:Lipschitz} indicates that the frequency response $h({\bm \l}, \om)$ does not change faster than linear since we have
\begin{equation*}
\begin{split}
    |h({\bm \l}_1, \om) - h({\bm \l}_2, \om)| & = |{\bf \nabla}_{\bm \l}^T h({\bm \l}_{1,2}, \om) \cdot ({\bm \l}_1 - {\bm \l}_2)| \\
    & \leq \|{\bf \nabla}_{\bm \l}^T h({\bm \l}_{1,2}, \om)\|_2 \cdot \|{\bm \l}_1 - {\bm \l}_2\|_2\\
    & \leq C_L \|{\bm \l}_1 - {\bm \l}_2\|_2.
\end{split}
\end{equation*}
The first inequality follows from triangle inequality while the first and third lines are direct application of Equations \eqref{eq:grad} and \eqref{eq:Lipschitz} respectively.
On the other hand, the condition in \eqref{eq:Lipschitz2} states that the variability in the frequency response should be low at higher eigenvalues. In other words, filters that meet \eqref{eq:Lipschitz2}, albeit stable, have a near-flat response at higher frequencies and therefore have poor discriminibilty between high graph frequencies. 
% This matches an observation discussed in previous works, e.g., \cite{Gama2020}, about the trade-offs between stability and discriminibility of graph filters. 
Under the two conditions, Theorem \ref{thm:STGF} characterizes the stability properties of STGFs. 
\begin{theorem*}[STGF Stability]\label{thm:STGF}
Consider an STGF with coefficients $\{h_k\}_{k=0}^K$ running over a sequence of $K$ identical GSOs ${\bf S}$ with output $\bf Y$ as in \eqref{eq:STGF}. Consider a sequence of $K$ realizations of $RES(\cG, p)$ with GSOs $\{{\bf S}_k\}_{k=1}^K$ under which the filter output $\tilde{\bf Y}$ is captured in \eqref{eq:stochSTGF}. Let ${\bf S}_k = {\bf S} + {\bf E}_k$, for $k = 1, \dots, K$, and ${\bf S}_0 = {\bf I}$.
If Assumption \ref{ass:Lipschitz} holds, the expected difference between the filter outputs satisfies
\begin{equation}
    \mathbb{E}[\| \tilde{\bf Y} - {\bf Y}\|_F^2 ] \leq C(1-p)\|{\bf X}\|_F^2 + \cO \left((1-p)^2\right),
\end{equation}
where $C= \a N C_L^2$, and the scalar $\a$ is either the maximum node degree if $\bf S$ is the adjacency or $2$ if it is the Laplacian.
\end{theorem*}
Theorem 1 affirms that the bound depends on some graph and filter parameters as well as the probability of edge drop $1-p$. It comes with no surprise that $N$ and $1-p$ contribute to the stability constant since holding either of them fixed and increasing the other leads to dropping more edges while constructing the random graph realization ${\cal G}_k$. Thus the distance between the GSOs $\bf S$ and ${\bf S}_k$ increases and affects the stability of the model. We also observe that filters with lower $C_L$ are more robust to perturbations. The constant $C_L$ describes the variability in the filter response (cf. \eqref{eq:Lipschitz}) and smaller values of $C_L$ indicates that the filter response changes slowly. This implies that the filter smoothness is key to provoke stability.

To analyze the stability of ST-GNNs, we consider a class of nonlinearities $\sigma$ that is Lipschitz continuous as stated in Assumption \ref{ass:nonlinearity}.

\begin{assumption*} \label{ass:nonlinearity}
    The nonlinearities $\s$ are Lipschitz-continuous functions with a Lipschitz constant $C_{\s} > 0$ such that, for any two vectors ${\bf x}_1$ and ${\bf x}_2$, we have
    \begin{equation}
        \label{eq:non-linear}
        \|\s({\bf x}_1) - \s({\bf x}_2)\|_2 \leq C_{\s} \|{\bf x}_1 - {\bf x}_2\|_2.
    \end{equation}
\end{assumption*}
This assumption is satisfied by the activation functions usually used in GNNs, e.g., ReLU and hyperbolic tangent (tanh) functions. When Assumptions \ref{ass:Lipschitz} and \ref{ass:nonlinearity} hold, the stability of ST-GNNs is determined by Theorem \ref{thm:STGNN}.
\begin{theorem*}[ST-GNN Stability]\label{thm:STGNN}
Consider an L-layer ST-GNN ${\bf \Phi}({\bf X}; {\bf S}, \cH)$ with $F$ features per layer and $F$ input and output features.
Let the STGF with coefficients $\{h_{kl}^{fg}\}_{kl}^{fg}$ at each layer satisfy Assumption \ref{ass:Lipschitz} and have a unit norm, i.e., $\left| h({\bm \l}, \om) \right| \leq 1$. Also, consider the same ST-GNN runs over a sequence of $K$ realizations of $RES(\cG, p)$ with output $\tilde{\bf \Phi}({\bf X}; {\bf S}, \cH)$. If the nonlinear activation functions satisfy Assumption \ref{ass:nonlinearity}, the expected difference between the outputs is bounded by
\begin{equation}
\begin{split}
    \mathbb{E} \left[\|\tilde{\bf \Phi}({\bf X}; {\bf S}, \cH) - {\bf \Phi}({\bf X}; {\bf S}, \cH)\|_F^2 \right] & \\ \leq  C (1-p)  \|{\bf X}\|_F^2 + & \cO\left( (1-p)^2 \right),
\end{split}
\end{equation}
where $C = \a N L^2 C_L^2 C_\s^{2L} F^{2L}$ with $\a$ being the same constant in Theorem \ref{thm:STGF}.
\end{theorem*}
Besides all the factors that control the stability of STGFs, Theorem \ref{thm:STGNN} includes some design factors, e.g., the number of features $F$, the number of layers $L$, and the Lipschitz constant of the nonlinearities $C_\sigma$. It is worth noting, however, that the bound grows exponentially with the number of layers, which suggests that deep networks are less robust to perturbations. This result is in line with other works that recommend the use of GNNs with only a few layers \cite{chen22}. The authors in \cite{wu19} even show that a single-layer GNN with multi-tap filters has a performance that is comparable, or even superior, to that of multi-layer single-tap GNNs. For our considered tasks, we have also observed that a small number of layers is sufficient to ensure stability without sacrificing performance/expressivity. Detailed study of the expressive power of GNNs can be found in \cite{kanatsoulis2022}.

%% file: Sections/Simulations.tex
We consider the flocking problem in \cite{tolstaya20a}, where the objective is to coordinate a swarm of agents to move together at the same velocity and avoid collisions. The problem has a closed-form optimal solution (see \cite{Tanner2003}). However, this solution requires access to data from all the agents and therefore does not allow decentralized settings. On the other hand, solutions that rely on gathering information from local neighborhood are harder to find. ST-GNNs can close this gap by learning a decentralized policy using imitation learning.

To construct a dataset, each agent is associated with a time sequence of positions $\mathbf{p}_i \in \mathbb{R}^{2 \times T}$ and velocities $\mathbf{v}_i \in \mathbf{R}^{2 \times T}$ as node features. The agent is also associated with optimal accelerations $\mathbf{u}_i \in \mathbb{R}^{2 \times T}$ as node predictions, which are pre-computed with a centralized algorithm. The node features are computed at each time step  $t$ via the system dynamics as soon as the model predicts the acceleration $\hat{\bf u}_i(t-1)$, as described in \cite{tolstaya20a}. The agents form a communication network that varies over time. The underlying graph ${\cal G}(t)$ contains an edge between nodes $i$ and $j$ if and only if they are within the communication range of each other, i.e., $\|{\bf p}_i(t) - {\bf p}_j(t)\|_2 \leq R$. Since the graph depends on the present positions of the agents, we cannot forecast the graph sequence before training/execution.
For our experiments, the dataset contains $400$ examples for  training, $80$ for validation, and another $80$ for testing. 

We train an ST-GNN that consists of one layer with $K=3$, $F=64$ and $\tanh$ activation function. We then implement a local layer at each node to predict the agent accelerations $\hat{\bf u}_i(t)$. We use ADAM \cite{KingmaB14} with learning  rate $5 \times 10^{-4}$ and forgetting factors  $\beta_1 = 0.9$ and $\beta_2 = 0.999$ in order to optimize the MSE between the optimal ${\bf u}_i(t)$ and the predicted $\hat{\bf u}_i(t)$. We train our model over $30$ epochs and used validation input-output pairs to choose the one that minimizes the velocity variation among the agents,
\begin{equation*}
    \text{cost } = \sum_{i=1}^N \left\|{\bf v}_i - \frac{1}{N} \sum_{j=1}^N {\bf v}_j \right\|_F^2 ,
\end{equation*}
 averaged over the validation dataset. 
 
 \begin{figure}[t!]
    \centering
    \includegraphics[width=\columnwidth]{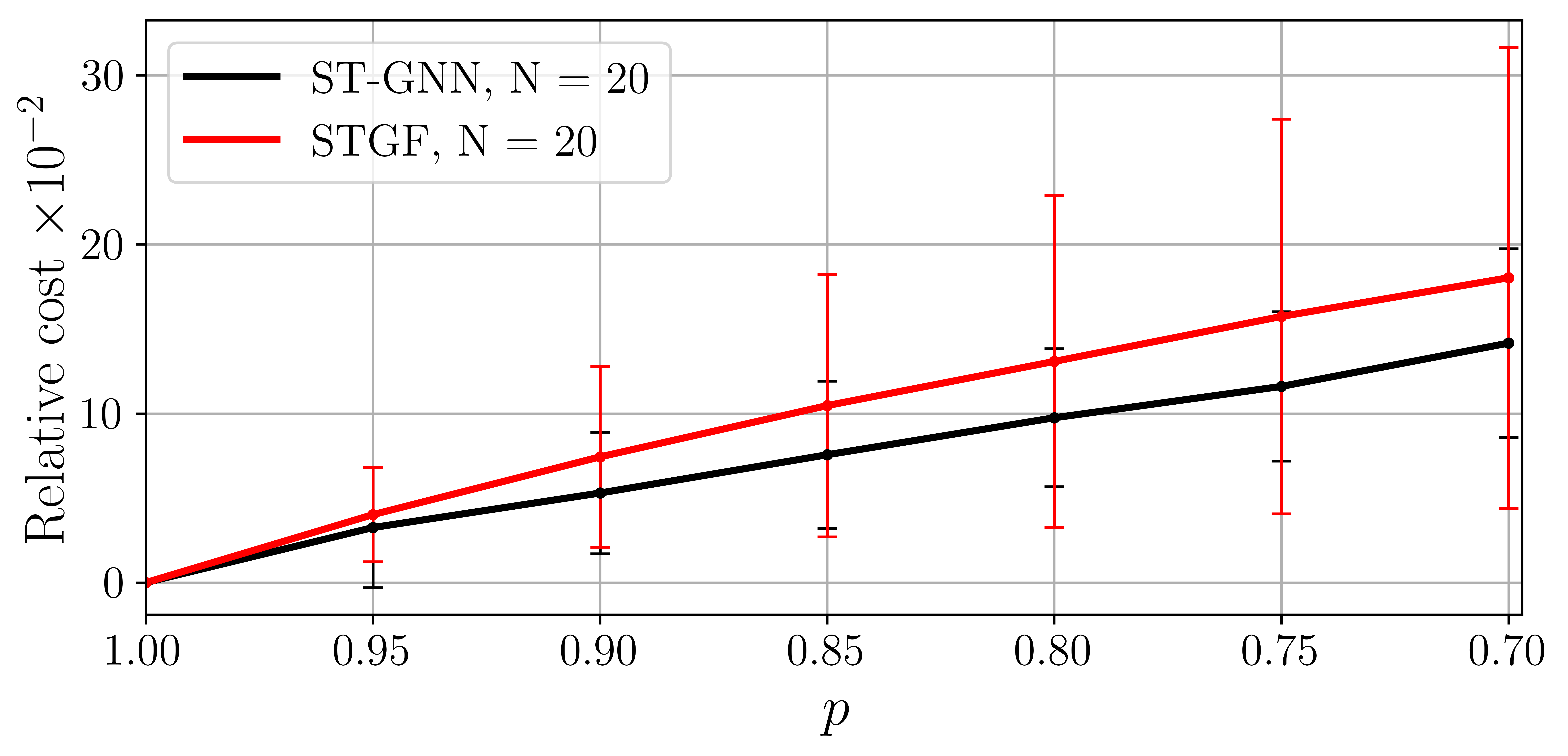}
    \includegraphics[width=\columnwidth]{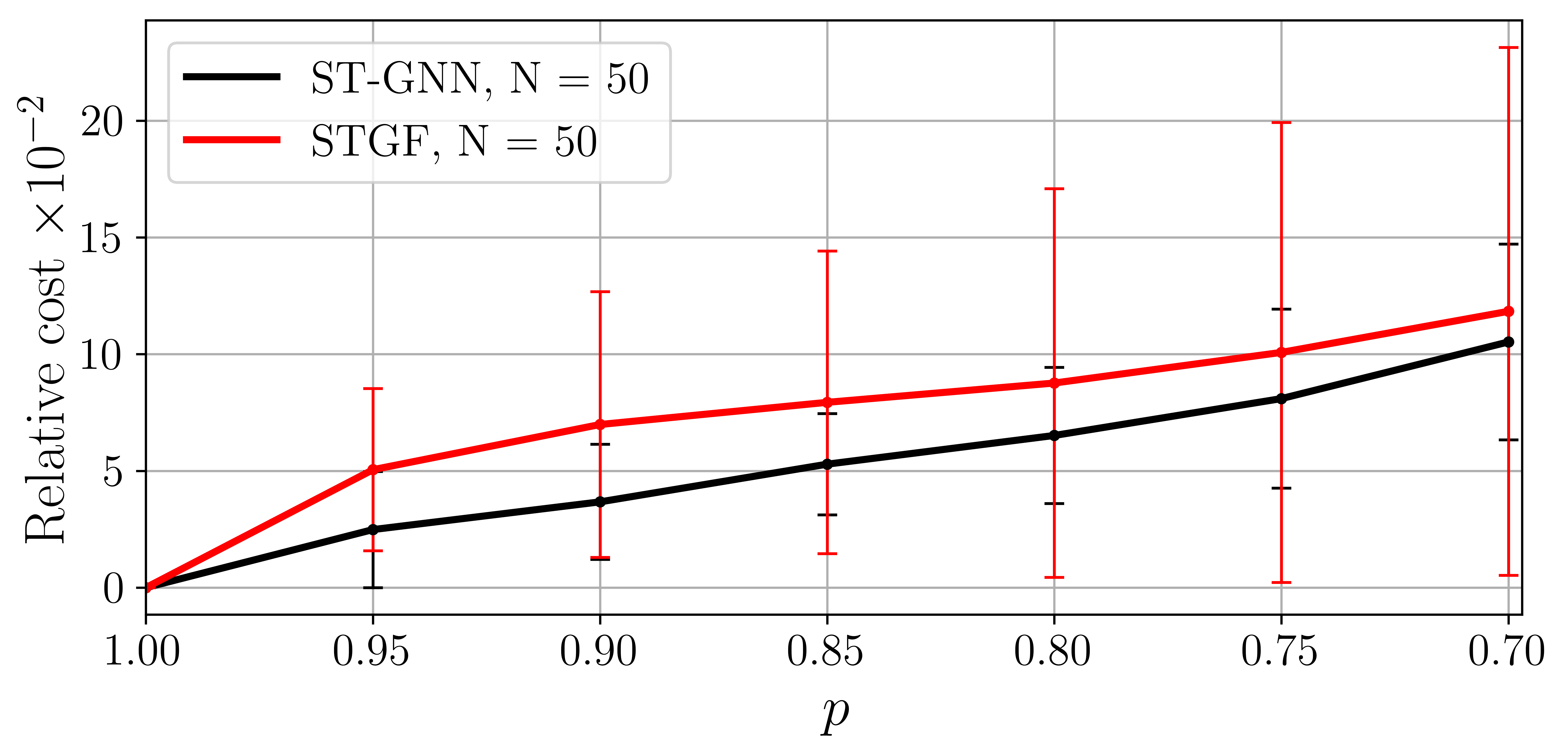}
    \includegraphics[width=\columnwidth]{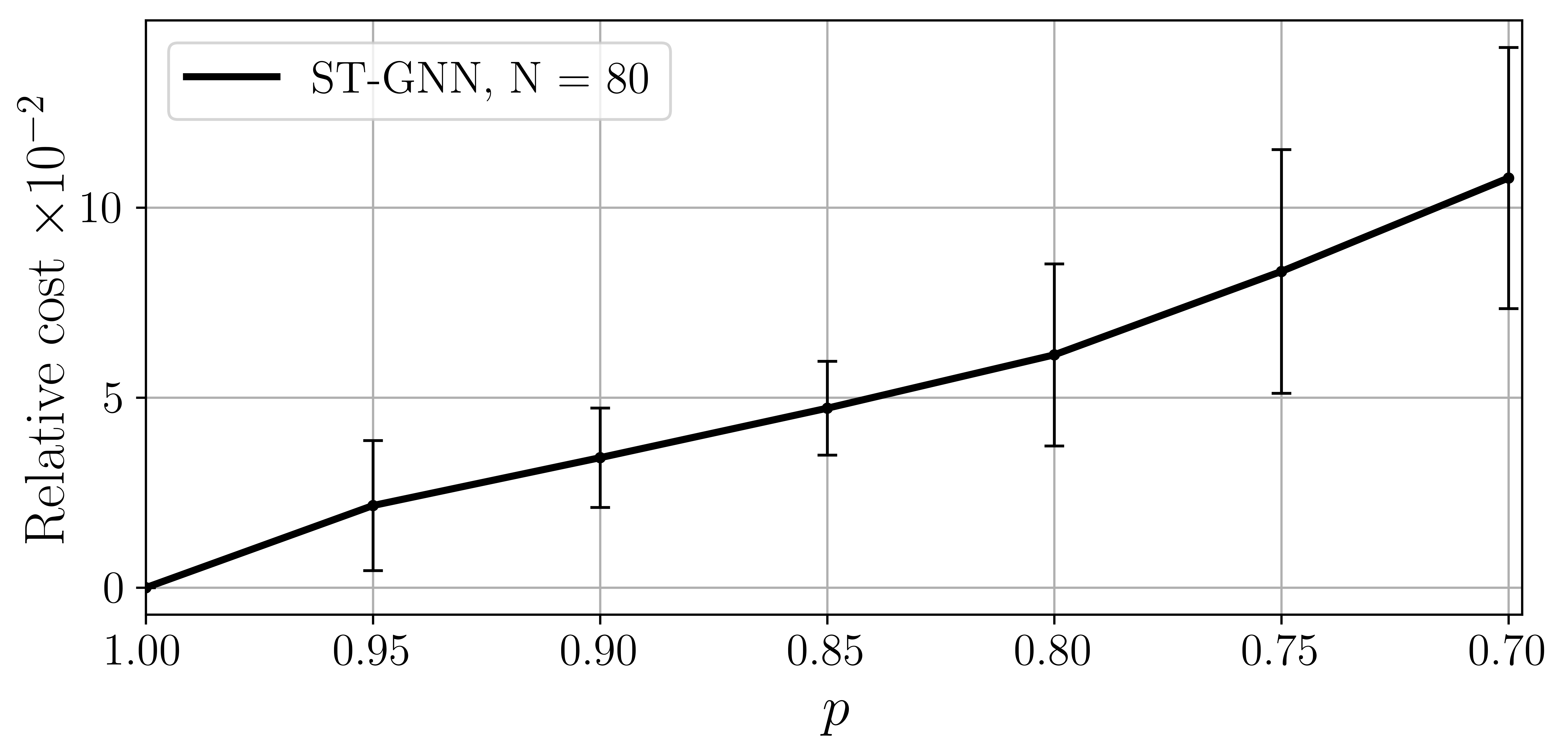}
    \caption{Relative cost induced by stochastic perturbations with different sampling probabilities $p$ for $N=20$ (top), $N=50$ (middle), and $N=80$ (bottom).}
    \label{fig:1}
\end{figure}
 
 In our first experiment, we train an ST-GNN on a fixed graph and test it over a sequence of random graphs sampled according to definition \ref{def:RES}. Contrary to \cite{tolstaya20a}, we train our model on the average of the graph sequence. But since we cannot forecast the graphs before training as we mentioned before, we use the agent's positions computed with the optimal accelerations to compute the graph sequence and, in turn, their average. During execution, we sample a graph sequence from the average graph with a sampling probability $p \in \{1, 0.95, 0.9, 0.85, 0.8, 0.75, 0.7\}$. Note that the sampled graph sequences no longer represent the agent's proximity since we decide on the fixed graph before training and we sample randomly from it at execution. Fig. \ref{fig:1} shows that the relative cost increases linearly with the edge-drop probability $1-p$, matching the result of Theorem \ref{thm:STGNN}. We also train an STGF with $K=4$ and $F=32$ and verify the linear bound of Theorem \ref{thm:STGF}. We, however, observe that ST-GNNs show less relative cost compared to STGFs even though we used a lower number of features with the latter. We attribute this observation to the $\tanh$ activation function being a nonexpansive mapping (i.e., we have $C_\sigma \leq 1$).
 
\begin{figure}[t!]
    \vspace{-0.6cm}
     \centering
    \begin{subfigure}{0.49\columnwidth}
        \centering
        \includegraphics[width=1.15\columnwidth]{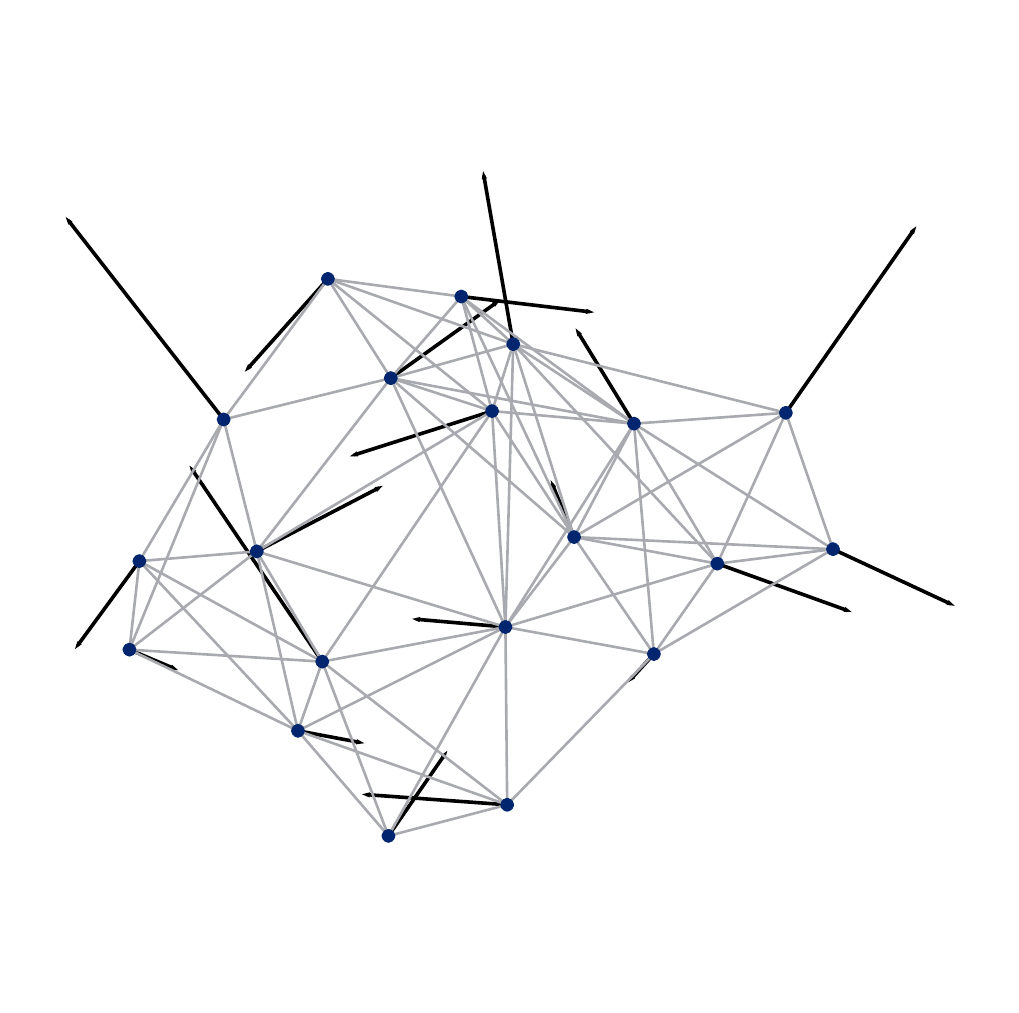}\\ \vspace{-0.43cm}
        \includegraphics[width=1.1\columnwidth]{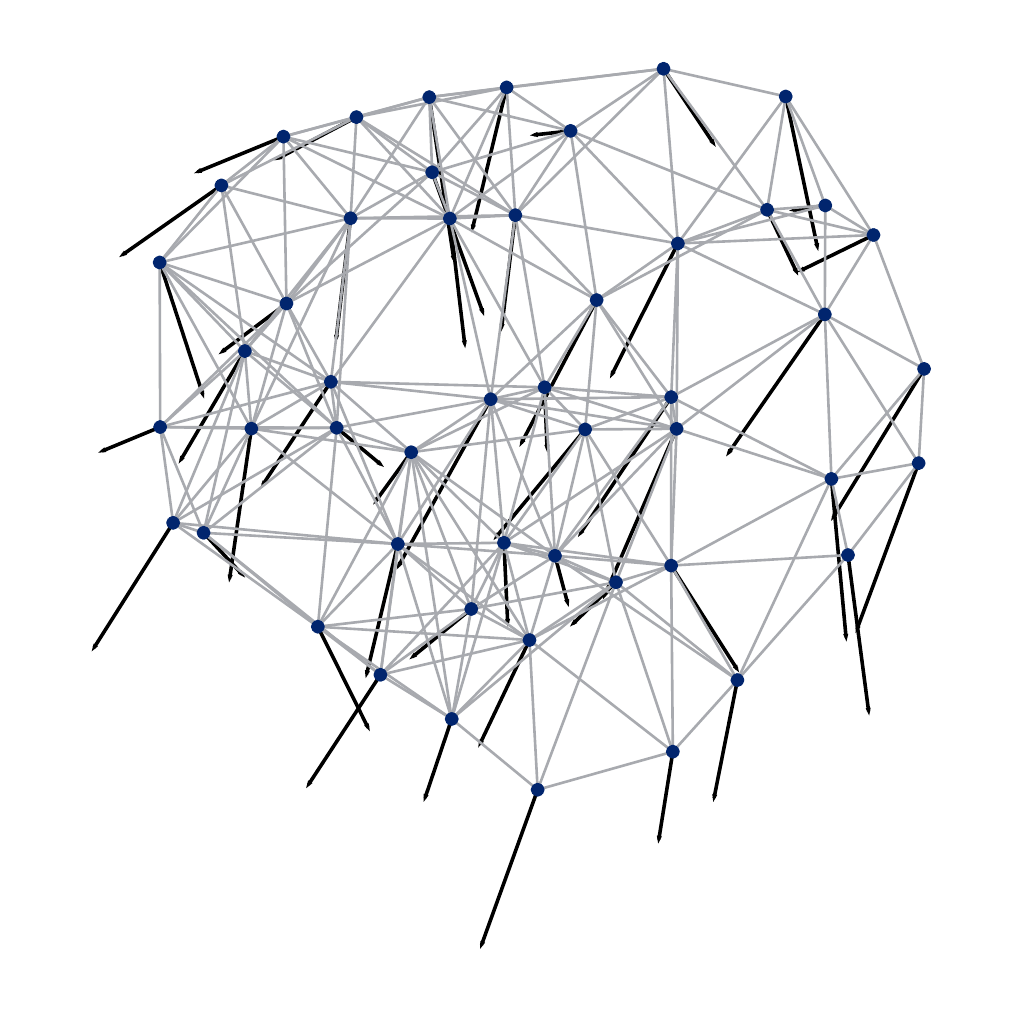}
        \vspace{-0.8cm}
        \caption{$t=0$s}
    \end{subfigure}
    \begin{subfigure}{0.49\columnwidth}
        \centering
        \includegraphics[width=\columnwidth]{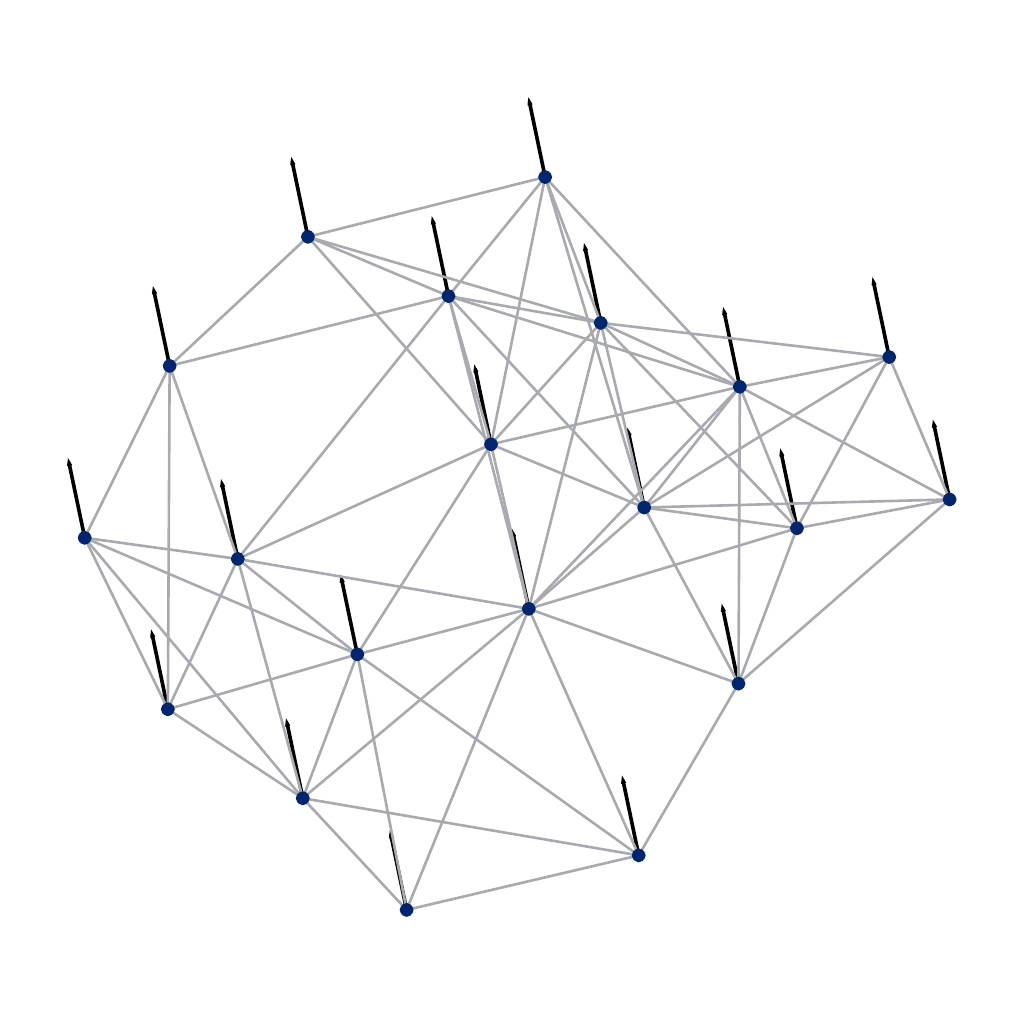}\\ \vspace{-0.43cm}
        \includegraphics[width=1.1\columnwidth]{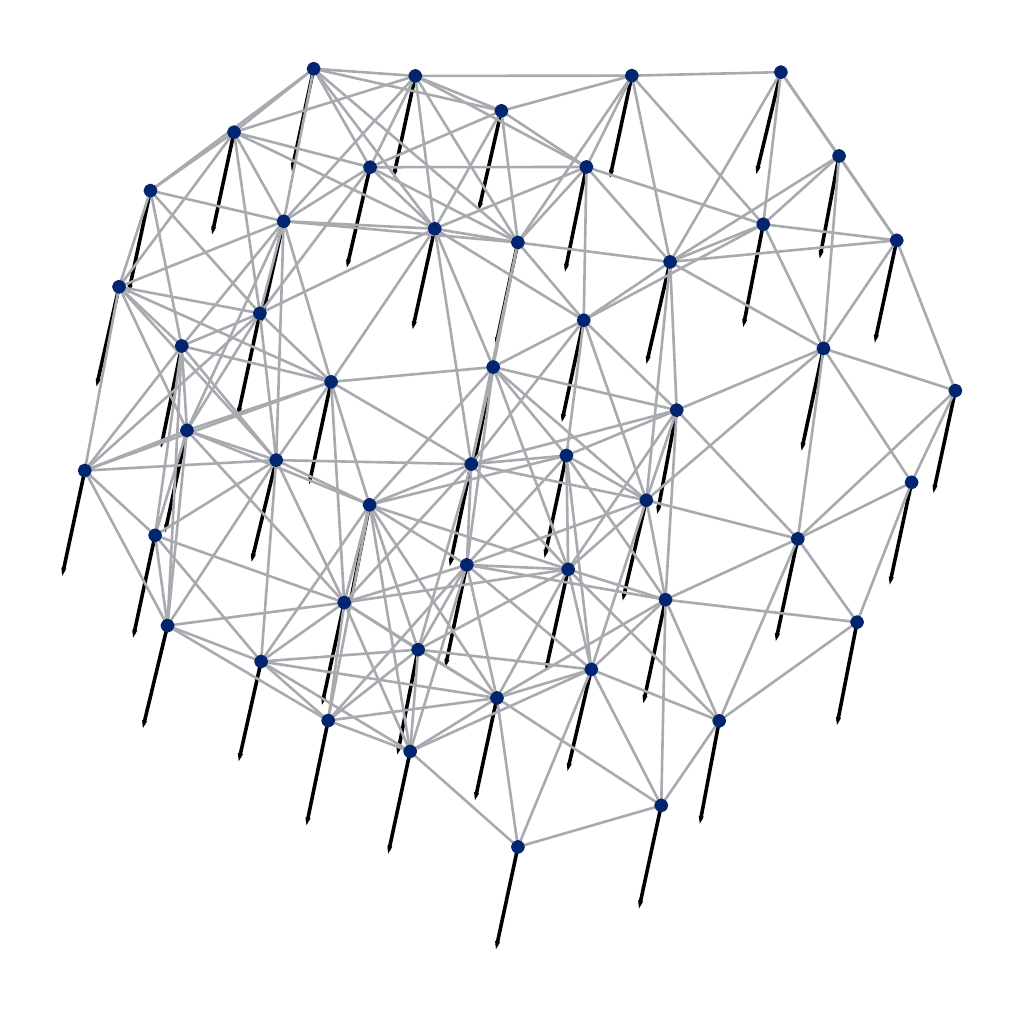}
        \vspace{-0.8cm}
        \caption{$t=1.99$s}
    \end{subfigure}
    \caption{Snapshots of two test examples: (top) $N=20$ and (bottom) $N=50$. The dots represent the agents, the gray lines illustrate the underlying graphs, and the dark arrows show the agents' velocity.}
    \label{fig:2}
\end{figure}

 We train a generalized ST-GNN, where the graphs are formed according to the communication networks between the agents. The graphs are constructed based on a communication range of $R=2m$ as described at the beginning of this section and as in \cite{tolstaya20a}. Thus the graphs are time varying during both training and execution. Fig. \ref{fig:2} illustrates snapshots of the output of a generalized ST-GNN for two test examples with unseen time-varying graphs of sizes $N=20$ and $N=50$ nodes. The snapshots were taken at the beginning and end of a time interval of $2$s. We observe that all the agents managed to move with the same velocity and in the same direction by the end of the time interval. This indicates that the trained ST-GNN is successfully transferred to unseen graphs, which validate our theoretical results. 

%% file: Sections/Conclusions.tex
The paper studied the stability of ST-GNNs to stochastic graph perturbations and showed that we can transfer learned models over time-varying graphs. Our analysis also enabled us to design generalized ST-GNNs that can be trained over time-varying graphs. We concluded that the stability bound is controlled by some graph, filter, and design parameters. The bound increases linearly with the graph size as well as the size of graph perturbations, represented by the edge-drop probability. It also depends on the Lipschitz constant of STGFs and the nonlinearities. 
Among all those parameters, only the Lipschitz constant of the filters cannot be decided before training since the filter is learnable. The stability can though be enhanced by forcing the filter to have a low Lipschitz constant during training, which is left as a future work.